\documentclass{article}

\usepackage{microtype}
\usepackage{graphicx}
\usepackage{xspace}
\usepackage{booktabs} 

\usepackage{hyperref}
\usepackage{float}
\usepackage{subfig}
\usepackage{tabularx}

\usepackage{amsmath,multicol}
\usepackage{color, colortbl}
\usepackage{xcolor}

\usepackage{listings}  

\usepackage{multirow}
\usepackage{makecell}

\usepackage{tcolorbox}
\usepackage{enumitem}

\definecolor{lightgray}{gray}{0.5}

\newcommand{\llm}{\texttt{PhoneLM}\xspace}

\newenvironment{nscenter}
{\parskip=5pt\par\nopagebreak\centering}
 {\par\noindent\ignorespacesafterend}

\usepackage[accepted]{icml2021}

\icmltitlerunning{\llm: an Efficient and Capable Small Language Model Family}

\begin{document}

\twocolumn[
\icmltitle{\llm: an Efficient and Capable Small Language Model Family\\through Principled Pre-training}




\begin{icmlauthorlist}
\icmlauthor{Rongjie Yi}{to}
\icmlauthor{Xiang Li}{to}
\icmlauthor{Weikai Xie}{to}
\icmlauthor{Zhenyan Lu}{to}
\icmlauthor{Chenghua Wang}{to}
\icmlauthor{Ao Zhou}{to}
\icmlauthor{Shangguang Wang}{to}
\icmlauthor{Xiwen Zhang}{goo}
\icmlauthor{Mengwei Xu}{to}
\end{icmlauthorlist}

\icmlaffiliation{to}{Beijing University of Posts and Telecommunications (BUPT), China}
\icmlaffiliation{goo}{Helixon Research}

\icmlcorrespondingauthor{Mengwei Xu}{mwx@bupt.edu.cn}


\vskip 0.3in
]



\printAffiliations{} 

\begin{abstract}
The interest in developing small language models (SLM) for on-device deployment is fast growing.
However, the existing SLM design hardly considers the device hardware characteristics.
Instead, this work presents a simple yet effective principle for SLM design: architecture searching for (near-)optimal runtime efficiency before pre-training.
Guided by this principle, we develop \llm SLM family (currently with 0.5B and 1.5B versions), that acheive the state-of-the-art capability-efficiency tradeoff among those with similar parameter size.
We fully open-source the code, weights, and training datasets of \llm for reproducibility and transparency, including both base and instructed versions.
We also release a finetuned version of \llm capable of accurate Android Intent invocation, and an end-to-end Android demo.
All materials are available at \url{https://github.com/UbiquitousLearning/PhoneLM}.


    
\end{abstract}
\section{Introduction}

In last few years, the striking progress has been made in large language models, attributed to the scaling-up ability of transformer.
One the other hand, we also notice growing interests in small language models (SLMs), which typically encompass sub- or a few billions of parameters and facilitate on-device deployments~\cite{lu2024small,yuan2024mobile}.
In practice, SLMs have been shipped to commercial off-the-shelf devices on a vast scale.
For instance, the latest Google/Samsung phones have built-in LLM service (Gemini Nano), through which third-party mobile apps can freely enjoy LLM capability through text prompts or LoRA~\cite{hu2021lora}.
Apple also introduces SLMs to facilitate privacy-preserving on-device intelligence tasks such as refining text and prioritizing notifications in iOS~\cite{apple-ios}.

\begin{figure}[t]
	\centering
	\subfloat[Chat]{
		\includegraphics[width=0.48\linewidth]{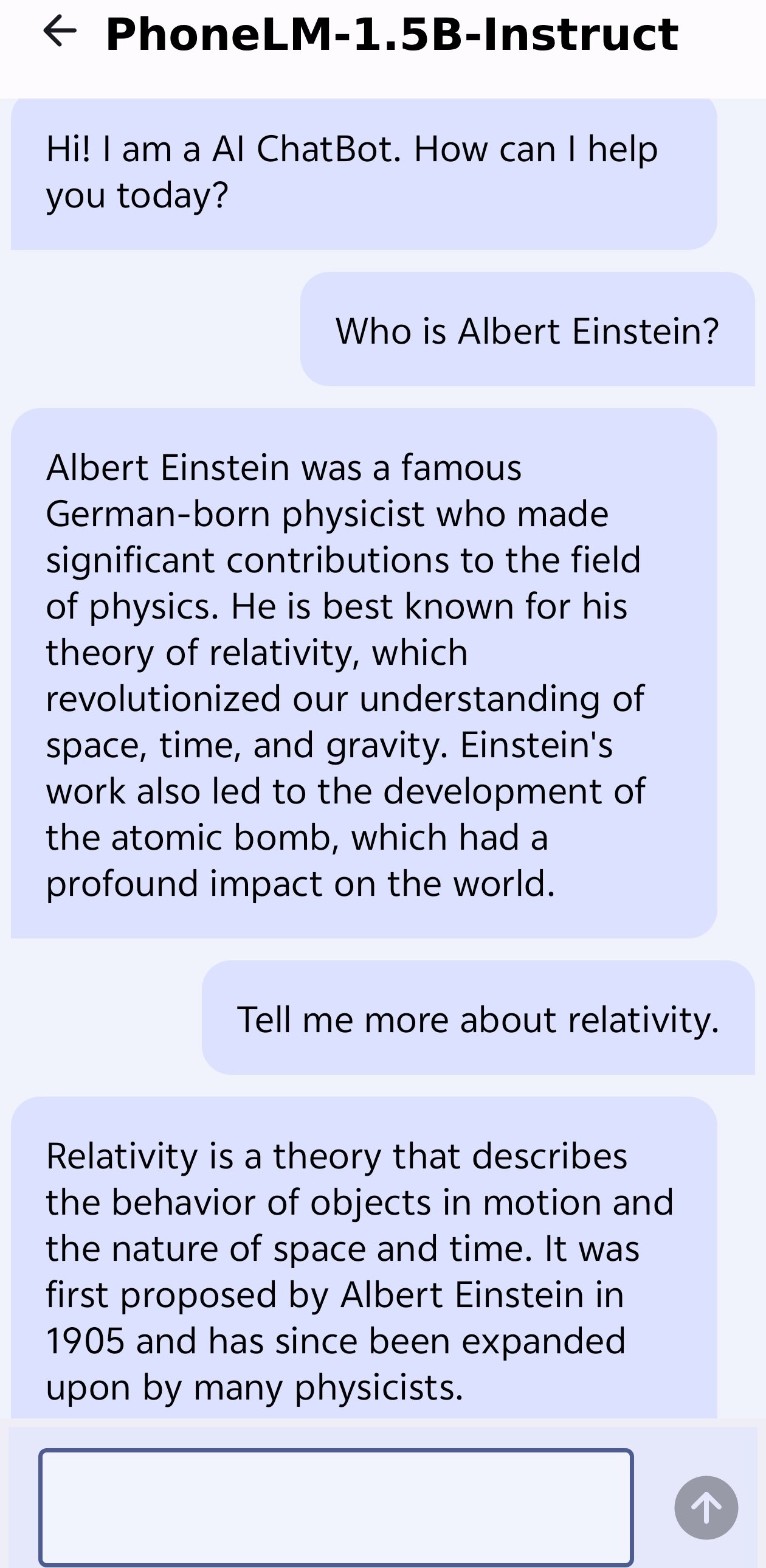}}
	\subfloat[Android Intent Invocation]{
		\includegraphics[width=0.49\linewidth]{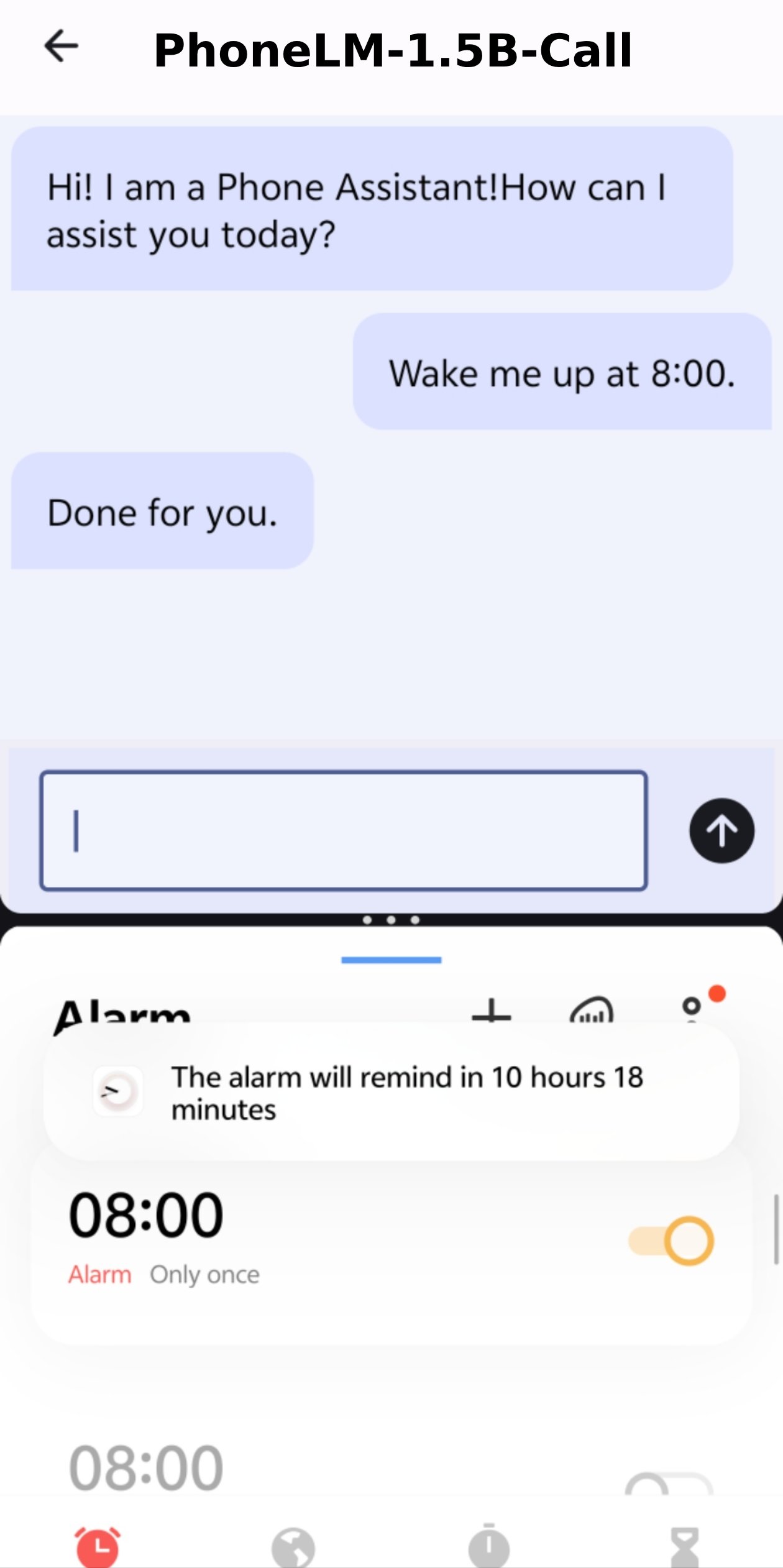}}
	\caption{An end-to-end Android demo of \llm's capability. 
        (a) shows an example of a user having a conversation with \llm-1.5B-Instruct; 
        (b) shows an example of a user invokes an Android intent through chatting with \llm-1.5B-Call.
        }
	\label{fig:experimental-innvaction}
\end{figure}

On-device SLM deployment is extremely challenging due to the resource scarce of edge devices~\cite{xu2024survey}.
While there has been plenty of open-sourced SLMs, e.g., Microsoft Phi family~\cite{microsoft_phi_3_mini}, that are claimed to be designed for resource-constrained devices, we found rare evidences supporting it except its relatively small parameter size.
Motivated by the absence of a high-level principle for SLM design, we ask a question: \textit{beyond using a small parameter size, what else can model developers do to better support on-device deployment with limited resources?}

In this work, we propose an intuitive yet effective principle for constructing on-device small language models:
\textbf{searching for an resource-efficient architecture on a given hardware before pretraining}.
It fundamentally differs from traditional SLM pipeline in that it moves the consideration of resource efficiency ahead of pre-training, while existing practice typically puts performance optimizations after pre-training (e.g., PTQ) but searches for an architecture with best capability (e.g., through observations on loss curve)~\cite{hu2024minicpm}.
The principle is reasoned with two observations.
(1) According to the scaling law~\cite{kaplan2020scaling}, the final model accuracy is not sensitive to the model configurations in a wide range; yet our experiments in $\S$\ref{sec:motivation} demonstrate the opposite finding for inference speed, where the same-sized SLMs (1.5B) can run with up to 3.13$\times$ speed gap (compared with OPT-1.3B) on the same smartphone.
(2) The cost of pre-training SLMs for different devices will be amortized by deploying SLM as a system-level service that delivers language ability to third-party apps, e.g., Google AICore~\cite{google-aicore}.
In such circumstance, the pre-training cost of SLMs for each device is one shot, regardless of how many applications it serves~\cite{yin2024llm}.

Guided by this principle, we develop and release \llm for smartphone hardware (e.g., Qualcomm Snapdragon SoC), a family of pre-trained and instructed SLMs.
It now includes 5 model variants: \llm-0.5B, \llm-0.5B-Instruct, \llm-1.5B, \llm-1.5B-Instruct, and \llm-1.5B-Call.
The first two are base models, while the other three are finetuned for instruction following and system-level function call in Android.
We also provide a few quantized versions to facilitate fast deployments.

There are three notable features of \llm:

First, \llm is extremely efficient through exhaustive ahead-of-pretraining architecture search on smartphone hardware.
For instance, \llm-1.5B runs at 58 tokens/second on Xiaomi 14 (Snapdragon 8Gen3 SoC) CPU, which is 1.2$\times$ faster than StableLM 2 1.6B and 1.6$\times$ faster than SmolLM 1.7B with similar parameter size.
The prefilling speed of \llm-1.5B even achieves 654 tokens/second on Xiaomi 14 NPU.
The underlying architecture of \llm is against recent SLM designs that converge to using SiLU (\llm adopts ReLU)~\cite{elfwing2018sigmoid} and a width-height ratio between 54.6--88.6 (\llm uses 134.7).
Such architecture not only offers speed advantage on CPU, but also facilitates the NPU-friendly quantization~\cite{xu2024empowering} and sparse activation~\cite{liu2023deja}.

\begin{figure*}[ht]
    \vskip 0.2in
    \begin{center}
    \centerline{\includegraphics[width=2\columnwidth]{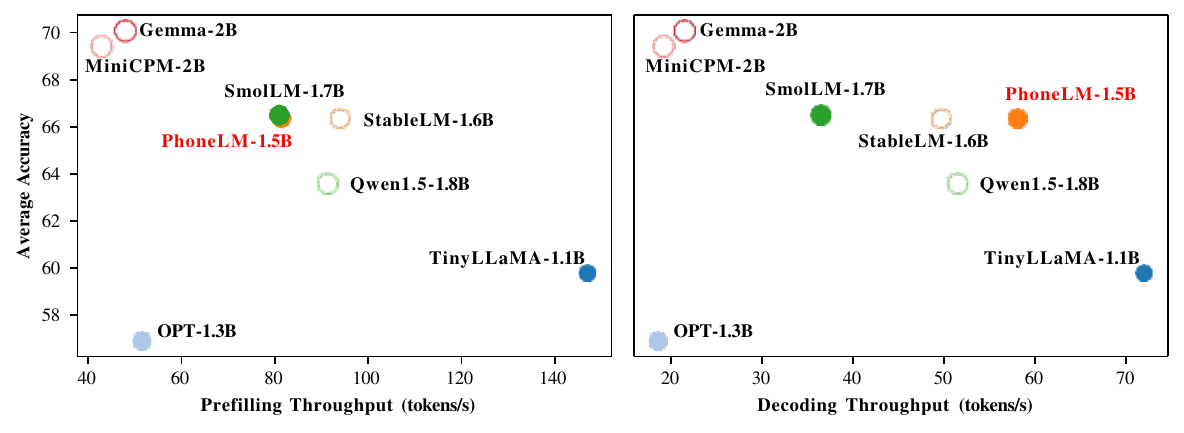}}
    \caption{The comparison of the average qccuracy and runtime performance between \llm-1.5B and SLMs with similar parameter quantities (1B to 2B). The average accuracy select seven NLP tasks to reflect the ability of the models (same as table~ref{tab:performance}), and the prefill/decode throughput is tested using the CPU on the Xiaomi 14 mobile phone. The closer the model is to the upper right corner, the better it is.
    Solid dots represent that the training data of the model is open source, and hollow dots represent that the training data of the model is closed source. 
    } 
    \label{fig:benckmark-throughput-matrics}
    \end{center}
    \vskip -0.2in
\end{figure*}

Second, \llm achieves impressive language capability with a small parameter size, as shown in Figure~\ref{fig:benckmark-throughput-matrics}.
Across 7 typical benchmarks (listed in Table~\ref{tab:performance}), \llm-1.5B scores 67.3\% accuracy on average, which is on par with the state-of-the-art SLMs with similar size trained on open datasets (i.e., SmolLM~\cite{smollm} 1.7B and DCLM~\cite{dclm} 1.4B).
It even achieves better capability than many SLMs trained on proprietary datasets such as Qwen 1.5 1.8B and StableLM 2 1.6B.
After finetuned, \llm-1.5B is also capable of having smooth conversations with humans, and controlling smartphones using Android intent through function calls.


Third, \llm is fully open-sourced, reproducible, and demonstrable.
\llm is trained on only open datasets without any manipulation.
We release the complete codebase to develop \llm, including the data preparation, training, fine-tuning, and evaluation procedures.
To showcase the capability of \llm in an end-to-end manner, we also release a demonstrable Android app powered by \llm and mllm~\cite{mllm} engine.
With the app, users can chat with \llm on devices or invoke OS function calls with human language, as shown in Figure~\ref{fig:experimental-innvaction}.

In a nutshell, \llm achieves the state-of-the-art speed-capability tradeoff for smartphones among the SLMs trained on open datasets.
We anticipate \llm, as well as the underlying principle of its development, to bring the community to the attentions on the importance of algorithm-hardware co-design and co-optimizations in SLMs.


\section{A Principle for SLM Development}
\label{sec:motivation}

\begin{figure*}[ht]
    \vskip 0.2in
    \begin{center}
    \centerline{\includegraphics[width=1.8\columnwidth]{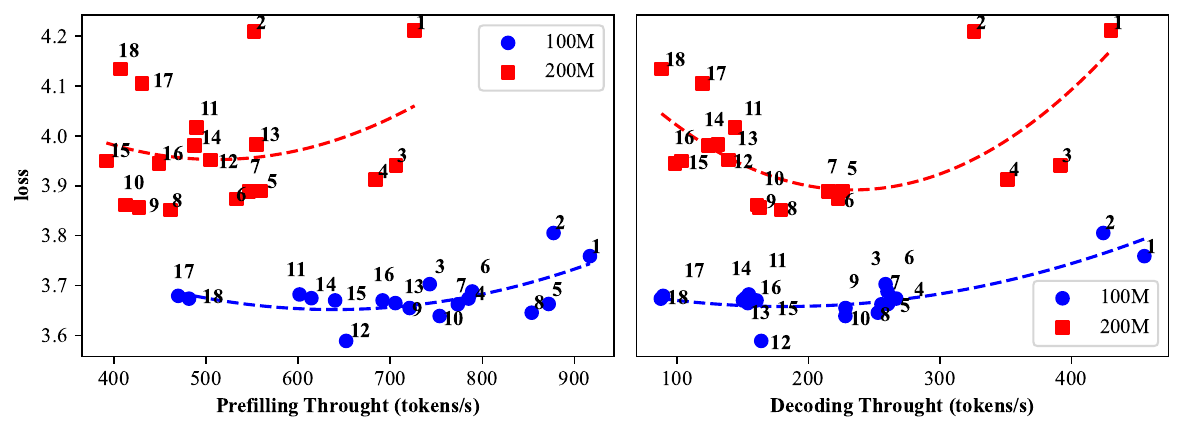}}
    \caption{The comparison of the throughput and ability of the models with parameter quantities of 100M and 200M. 
    More details of these model architecture are shown in appendix~\ref{sec:appendix-100200-setting}
    }
    \label{fig:motivation-100_200M-matrics}
    \end{center}
    \vskip -0.2in
\end{figure*}

\begin{nscenter}
\textit{SLM shall adapt to the target device hardware}.
\end{nscenter}
A key argument of this work is that, unlike on clouds, the SLM architecture and development shall adapt to the specific hardware for runtime efficiency as the first-class concern.
Throughout this work, the ``SLM architecture'' mainly refers to the hyperparameters of transformer-decoder models, including the types of attention (MHA, GQA, etc.), activation function of feed forward network (FFN), depth and width of the model, etc.

\textbf{Motivating experiments.}
To support the principle proposed, we test a bunch of SLMs with 100M and 200M parameters using various configurations on 2B tokens (dataset is the same as used to train \llm).
We then compare their loss on the same validation dataset.
At the same time, we tested the inference speeds of these models using the inference engine \textit{mllm}~\cite{mllm} on a smartphone equipped with the Snapdragon 8Gen3 SoC. 
The results of average metric (introduced in Section~\ref{sec:pretraining-results}) and inference speed (throughput) are shown in figure~\ref{fig:motivation-100_200M-matrics}. 
More details of these model architectures are shown in appendix~\ref{sec:appendix-100200-setting}.
We fit a quadratic curve to the loss of the 100M and 200M models when training on the same 2B tokens of data.
Overall, fewer transformer layers, a larger model hidden size, and more attention heads tend to have faster inference speeds.

A key observation is that \textit{ runtime speed is more sensitive to the SLM architecture than the loss.}
For a given model size, the range of its runtime speed is much wider than that of the loss.
Comparing the SLMs with different sizes (100M and 200M), there is significant overlap of inference speed, but hardly any overlap of loss.
In other words, a model with 200M parameters is consistently more capable than the one with 100M parameters, but does not always run slower on devices.
The speed gap could be as large as 5$\times$ under the same model size.
With more training tokens, the loss gap would even close up according to our experiments.





\textbf{A principle of SLM development.}
Based on the insights, we present an intuitive yet effective principle for SLM development:
search for the most efficient architecture on given hardware, then pre-train it on datasets with best quality and most quantity as possible.
This principle differs from existing approaches that uses model capability as the target metric in SLM architecture search~\cite{hu2024minicpm}, leaving runtime optimizations in post-training stages.

\section{\llm: Smartphone-native SLM Family}

Following the proposed principle, we developed and trained \llm, a smartphone-native SLM family.
It has the following notable features: 
(1) Good runtime performance and capability.
(2) Convenient for smartphone deployment and more suitable for model inference using NPU.

In this section, we present the architecture and training details of \llm.



\subsection{Architecture}\label{sec:pretraining-architecture}


\begin{table}[ht]

    \centering
    
    \begin{tabular}{lcc}
        \toprule
        Model Size & 0.5B & 1.5B \\
        \midrule
        Hidden size    & 1,024 & 2,560 \\
        Intermediate Hidden Size & 4,864 & 6,816 \\
        Heads    & 16 & 16 \\
        Layers    & 24 & 19  \\
        Vocab size     & 49,152 & 49,152 \\
        Context Len    & 2,048  & 2,048 \\
        Training Tokens & 1.1T & 1.5T \\
        \bottomrule
    \end{tabular}
    \caption{\llm hyperparameters and training settings. Notably, only \llm-1.5B is developed with ahead-of-pretraining architecture search.}
    
    \label{tab:pretraining-architecture}
    
\end{table}


\llm uses the vanilla transformer decoder architecture.
The details of the \llm parameters (0.5B and 1.5B versions) are shown in Table~\ref{tab:pretraining-architecture}.
\llm uses a context length of 2,048 and reuses the SmolLM's tokenizer ~\cite{smollm} with a vocabulary size of 49,152.
It uses RoPE embedding and multi-head attention mechanism;
within its feed-forward component, it uses Gated FFN (as in Llama~\cite{touvron2023llama} and Qwen~\cite{bai2023qwen}), RMSNorm, and ReLU activation.


\begin{table}[t]
    \centering
    \small
	\begin{tabular}{ccc|cc}
        \hline
        \textbf{hidden} & \textbf{intermediate} & \textbf{layers} & \makecell{\textbf{prefilling}\\\textbf{(tokens/s)}} &  \makecell{\textbf{decoding}\\\textbf{(tokens/s)}} \\
        \hline
        2048 & 12288 & 16 & 70.75 & 55.12 \\
        2560 & 7680 & 18 & 64.98 & 60.60 \\
        \textbf{2560} & \textbf{6816} & \textbf{19} & \textbf{81.47} & \textbf{58.08} \\
        2048 & 10240 & 19 & 68.52 & 54.48 \\
        1792 & 10752 & 21 & 65.42 & 50.18 \\
        2048 & 8192 & 22 & 67.10 & 54.04 \\
        1792 & 8960 & 25 & 63.29 & 48.63 \\
        \hline
    \end{tabular}
\caption{The throughput of models with multiple structures of 1.5B on the Xiaomi 14 CPU (Snapdragon 8Gen3).}
\label{tab:pretraining-nas-latency}
\end{table}

\textbf{Hardware-specific, ahead-of-pretraining hyperparameter search for runtime resource efficiency.} 
According to the observation in section~\ref{sec:motivation}, we selected the hyperparameters of \llm by exhaustively searching on smartphone hardware.
We utilized the observations in section~\ref{sec:motivation} to set up many model structures as follows: (1) The number of model layers ranges from 15 to 25. (2) Use MHA with 16 heads and GQA with 4 groups. (3) The activation function is ReLU. (4) The ratio of intermediate size to hidden size is between 2 and 5. 
We conducted inference speed tests for different architectures of \llm-1.5B on Xiaomi 14. 
The structural designs of these models are presented in the table~\ref{tab:pretraining-nas-latency}.
We selected the model with the fastest inference speed as the final structure of \llm, as shown in the table~\ref{tab:pretraining-architecture}.  

\textbf{Using ReLU as the activation.}
Unlike recent SLMs that use SiLU or GELU as the activation function, \llm uses a rather legacy activation type ReLU.
There are two reasons.
First, calculating the ReLU operator on a smartphone is more efficient than the SiLU operator commonly used in LLM, especially for NPUs that specialize in integer calculations.  
Second, ReLU brings more sparsity to the FFN structure, facilitating the use of coefficient calculation techniques for inference acceleration on smartphones~\cite{song2023powerinfer,alizadeh2023llm}.

\textbf{Pre-quantized positional embedding.} 
We use RoPE (Rotary Positional Embedding) to inject positional information into our model. 
In order to accelerate the quantization calculation on smartphones, we quantize the values of $\sin$ and $\cos$ of RoPE to INT8 because it is a fixed cosine function without outliers, this quantization introduces minimal loss in accuracy.
We obtain the quantized $\sin$ and $\cos$ functions through the following formula:
\begin{align*}
    \cos, \sin &= RotaryEmbedding(shape(value))\\
    \cos_{max}&=\max\limits_{i}|{\cos}_i| \\
    \sin_{max}&=\max\limits_{i}|{\sin}_i| \\
    \cos_{int8}&=\lfloor\frac{\cos}{\cos_{max}}\times127+\frac{1}{2}\rfloor \\
    \sin_{int8}&=\lfloor\frac{\sin}{\sin_{max}}\times127+\frac{1}{2}\rfloor
\end{align*}
During forward propagation, positional encoding is added to the query and key in the following way:
\begin{align*}
    \cos = &\cos_{int8}\times\frac{\cos_{max}}{127}\\
    \sin = &\sin_{int8}\times\frac{\sin_{max}}{127}\\
    query, key &= ApplyRotaryEmbed(query, key, \cos, \sin)
\end{align*}

\begin{table}[ht]
    \centering
    \scriptsize
	\subfloat[Stable Training Stage]{
    \begin{tabular}{l|l|r}
        \hline
        \textbf{type} & \textbf{dataset} & \textbf{token} \\
        \hline
        web\qquad\qquad\qquad& DCLM-baseline~\cite{li2024datacomplm}\qquad\qquad\qquad & 1.35T \\
        \hline
        code & StarCoderData~\cite{li2023starcoder} & 112.75B \\
        \hline
        \multirow{2}{*}{math} & OpenWebMath~\cite{paster2023openwebmath} & 13.25B \\
        & Dolma-algebraic~\cite{dolma} & 12.75B \\
        \hline
        academic & Dolma-arxiv~\cite{dolma} & 29B \\
        \hline
        \multicolumn{2}{c|}{\textbf{total}} & \textbf{1.5T} \\
        \hline
    \end{tabular}
	\label{tab:pretraining-data-stage1}
    }
	\quad 
	\subfloat[Decay Stage]{
    \begin{tabular}{l|l|r}
        \hline
        \textbf{type} & \textbf{dataset} & \textbf{token} \\
        \hline
        web & DCLM-baseline~\cite{li2024datacomplm} & 10B \\
        \hline
        \multirow{2}{*}{code} & StarCoderData~\cite{li2023starcoder} & 1.575B \\
         & The Stack Smol & 0.95B \\
        \hline
        \multirow{2}{*}{acadamic}  & Dolma-arxiv~\cite{dolma} & 2.325B \\
         & Dolma-pes2o~\cite{dolma} & 2.35B \\
        \hline
        math instruct & MathInstruct~\cite{yue2023mammoth} & 65.25M \\
        \hline
        \multirow{3}{*}{chat instruct}  & UltraChat~\cite{ding2023enhancing} & 1.775B \\
         & OpenAssistant 2~\cite{kopf2024openassistant} & 42.25M \\
         & OpenHermes~\cite{openhermes} & 77.25M \\
        \hline
        \multirow{4}{*}{code instruct} & Magicoder Evol Instruct~\cite{magicoder} & 30.25M \\
         & CommitPackFT~\cite{muennighoff2023octopack} & 0.35B \\
         & Magicoder OSS Instruct~\cite{MagicoderOSSInstruct75K} & 43.5M \\
         & SlimOrca~\cite{SlimOrca} & 209.75M \\
        \hline
        \multirow{2}{*}{\makecell{function calling\\instruct}} & APIGen~\cite{liu2024apigen} & 48.25M \\
         & Glaive Function Calling~\cite{glaive-function-calling} & 57.5M \\
        \hline
        \multicolumn{2}{c|}{\textbf{total}} & \textbf{20B} \\
        \hline
    \end{tabular}
	\label{tab:pretraining-data-stage2}
    }
	\quad 
    \subfloat[Instruct Turning Stage]{
    \begin{tabular}{l|l|r}
        \hline
        \textbf{type} &\textbf{dataset} & \textbf{token} \\
        \hline
        math instruct & MathInstruct~\cite{yue2023mammoth} & 65.25M \\
        \hline
        \multirow{3}{*}{chat instruct}  & UltraChat~\cite{ding2023enhancing} & 1.775B \\
         & OpenAssistant 2~\cite{kopf2024openassistant} & 42.25M \\
         & OpenHermes~\cite{openhermes} & 77.25M \\
        \hline
        \multirow{4}{*}{code instruct} & Magicoder Evol Instruct~\cite{magicoder} & 30.25M \\
         & CommitPackFT~\cite{muennighoff2023octopack} & 0.35B \\
         & Magicoder OSS Instruct~\cite{MagicoderOSSInstruct75K} & 43.5M \\
         & SlimOrca~\cite{SlimOrca} & 209.75M \\
        \hline
        \multicolumn{2}{c|}{\textbf{total}} & \textbf{2.59B} \\
        \hline
    \end{tabular}
    \label{tab:pretraining-data-sft}
    }
    \caption{The classification of the datasets used in each stage and the number of their tokens. The description of the datasets is in appendix~\ref{sec:appendix-dataset}.}
    \label{tab:pretraining-data}
\end{table}

\begin{figure*}[t] 
	\centering  
	\subfloat[0.5B]{
        \label{fig:pretraining-training-loss-0.5B}
		\includegraphics[width=0.5\linewidth]{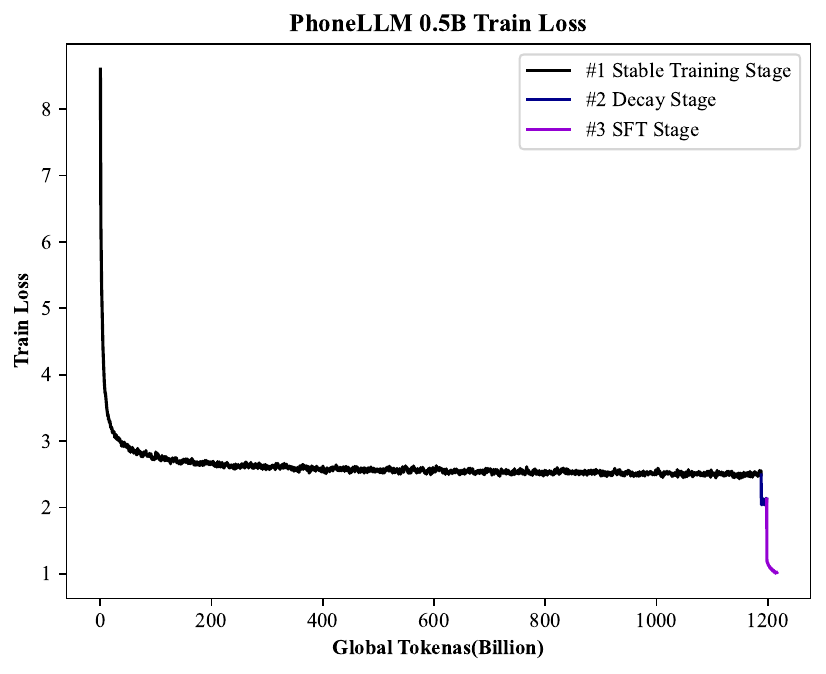}}
	\subfloat[1.5B]{
		\label{fig:pretraining-training-loss-1.5B}
		\includegraphics[width=0.5\linewidth]{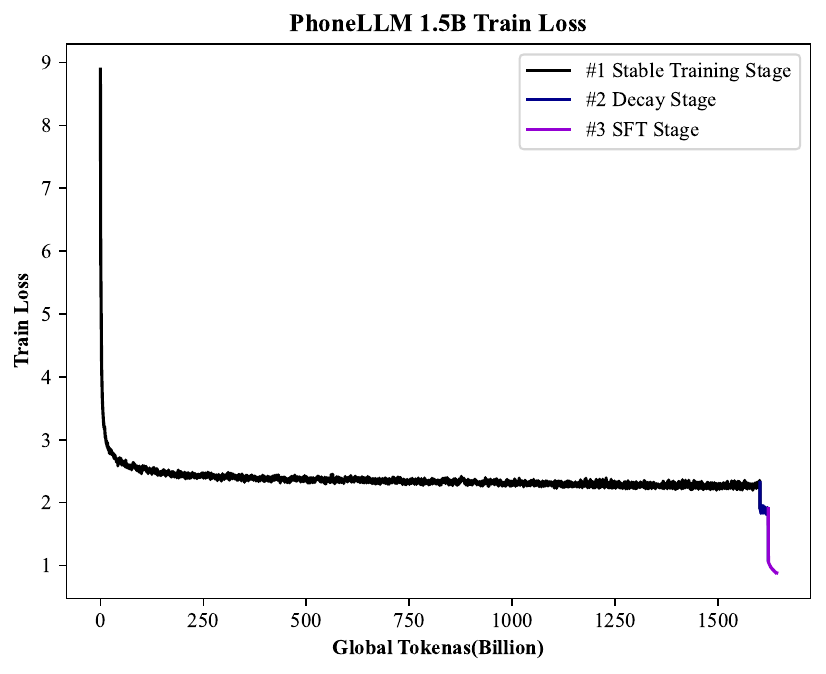}}
	\caption{Training loss}
	\label{fig:pretraining-training-loss}
\end{figure*}
\subsection{Pre-training}

The training of \llm has been set up as follows: 
(1) The optimizer is AdamW~\cite{loshchilov2017decoupled} with $\beta_1$ of 0.9, $\beta_2$ of 0.95, and $\epsilon$ of 1e-8.
(2) We use Fully Sharded Data Parallel (FSDP) to leverage multi-GPU and multi-node setups efficiently. 
(3) Another critical improvement is the integration of Flash Attention 2, an optimized attention mechanism. 
(4) We also use Zero Redundancy Optimizer(ZeRO), a memory optimization technique that reduces the models's memory footprint. 
(5) We use BF16 to accelerate the training process. 
The details of the setting of pre-training stage are shown in table~\ref{tab:pretraining-set}.

We use a dataset sourced from open datasets. 
For \llm-0.5B, we use 1.1 trillion tokens, and for \llm-1.5B, we use 1.5 trillion tokens.
In pre-training stage, we apply the weight decay, a learning rate warmup,  and a cosine learning rate decay schedule.

\llm is totally trained on open-sourced datasets without any manipulation, as shown in table~\ref{tab:pretraining-data}.
In the stable training stage, several open-source datasets are used, including DCLM-baseline, StarCoderData, OpenWebMath, Dolma.
The details of the training datasets are shown in appendix~\ref{sec:appendix-dataset}.
The pre-training loss of \llm family on the pretraining dataset is shown in figure~\ref{fig:pretraining-training-loss} with black line.

\subsection{Fine-tuning}
\label{finetune}

The fine-tuning of \llm base model is similar to MiniCPM~\cite{hu2024minicpm} and Llama 3~\cite{dubey2024llama}, which includes two stages: decay stage and Fine-tuning stage.
(1) Decay Stage.
We use a mixture of the pre-training data and high-quality supervised fine- tuning data, which is about 20 billion tokens.
In this stage, we use a linear learning rate decay schedule. 
(2) Fine-tuning Stage.
We find it still necessary to conduct a separate Fine-tuning stage. We utilize fine-tuning data similar to that in the decay phase but excludes pre-training data, totaling approximately 2.59 billion tokens. 
The learning rate for fine-tuning is set to match the final learning rate from the decay stage.
The optimizer in the Fine-tuning stage is the same as that in the pre-training stage for acceleration, but with different hyperparameter settings, which are shown in the table~\ref{tab:pretraining-set}.

\textbf{Instruct Tuning.}
In the decay stage, the data mixture contains some dataset from stable training stage, including DCLM-baseline, StarCoderData, and Dolma.
Then it contains some high-quality fine-tuning data,which is used in Fine-tuning stage.
The fine-tuning datasets are shown in table~\ref{tab:pretraining-data}, including APIGen, Stack Smol, UltraChat, MathInstruct, OpenAssistant 2, OpenHermes, CommitPackFT, OSS-Instruct, and SlimOrca.
The details of these datasets are shown in appendix~\ref{sec:appendix-dataset}.
The pre-training loss of Decay Stage and Fine-tuning Stage is shown in figure~\ref{fig:pretraining-training-loss}
Since we continue fine-tuning the model after the decay stage, the loss drops significantly at the beginning of each epoch.

\textbf{Function Call Tuning.}
To enhance the model's capability in smartphone operation, we fine-tuned the \llm on the DroidCall dataset, a synthetic dataset specifically focused on Android intent invocations generated by GPT-4.
The DroidCall dataset consists of 10k samples of function calling, encompassing simple, parallel, and nested call patterns. This dataset covers common Android operations, including setting alarms, configuring timers, composing email drafts, performing searches, and more.
We use LoRA to fine-tune \llm, adding adapter to all linear layers within both the attention layers and MLP layers
The fine-tuning process was configured with an initial learning rate of 1.41e-5, utilizing a rank (r) of 8 and an alpha value of 16. A linear learning rate scheduler was employed with a warmup ratio of 0.1. 
To ensure a minimal computational load and to increase inference speed, we used a minimalist prompt, which essentially only included function information and user queries.
 The final function calling model was derived from the optimal checkpoint of the fine-tuning process.
The details of prompt construction are shown in appendix~\ref{sec:appendix-function-calling}.

\begin{table}[t]
    \centering
	\subfloat[0.5B]{
        \begin{tabular}{|l|r|r|r|}
            \hline
            \textbf{stage} & \textbf{Stable} & \textbf{Decay} & \textbf{SFT} \\
            \hline
            Datasets (tokens) & 1.1TB & 20B & 2.59B \\
            \hline
            Learning Rate Scheduler & Cosine & Linear & None \\
            \hline
            Max Learning Rate & 4e-04 & 8e-05 & 4e-05 \\
            \hline
            Min Learning Rate & 8e-05 & 4e-05 & 4e-05 \\
            \hline
            Batch Size & 13.5M & 1.5M & 32M \\
            \hline
            Epoch & 1 & 1 & 7 \\
            \hline
            Training Days (A100) & 72$\times$10 & 16$\times$0.6 & 16$\times$1 \\
            \hline
        \end{tabular}
	\label{tab:pretraining-set-0.5B}
    }
	\quad 
	\subfloat[1.5B]{
        \begin{tabular}{|l|r|r|r|}
            \hline
            \textbf{stage} & \textbf{Stable} & \textbf{Decay} & \textbf{SFT} \\
            \hline
            Datasets (tokens) & 1.5TB & 20B & 2.59B \\
            \hline
            Learning Rate Scheduler & Cosine & Linear & None \\
            \hline
            Max Learning Rate & 4e-04 & 4e-05 & 2e-05 \\
            \hline
            Min Learning Rate & 4e-05 & 2e-05 & 2e-05 \\
            \hline
            Batch Size & 9M & 9M & 128M \\
            \hline
            Epoch & 1 & 1 & 8 \\
            \hline
            Training Days (A100) & 64$\times$35 & 64$\times$0.2  & 64$\times$1 \\
            \hline
        \end{tabular}
	\label{tab:pretraining-set-1.5B}
    }
    \caption{Training settings}
    \label{tab:pretraining-set}
\end{table}

\section{Experiment Results}

\begin{table*}[ht]
    \centering
    \scriptsize
    \subfloat[0.5B]{
        \begin{tabular}{l|c|c|c|ccccc|cc|c}
            \hline
            \textbf{Name} & \textbf{Size} & \textbf{Date} & \makecell{\textbf{Training}\\\textbf{tokens}} & \textbf{HellaSwag} & \textbf{WinoGrande} & \textbf{PIQA} & \textbf{SciQ} & \textbf{BoolQ} & \makecell{\textbf{ARC}\\\textbf{Easy}} & \makecell{\textbf{ARC}\\\textbf{Challenge}} & \textbf{Average} \\
            \hline
            Pythia~\cite{eleutherai_pythia_410m} & 410M & 23.03& 207B& 40.6 & 53.7 & 66.9 & 72.4 & 60.3 & 45.9 & 24.5 & 52.04 \\
            OPT~\cite{facebook_opt_350m} & 350M & 22.05&180B & 36.8 & 52.3 & 64.3 & 68.5 & 57.6 & 40.1 & 23.7 & 49.04 \\
            BLOOM ~\cite{bigscience_bloom_560m}& 560M & 22.11& 350B& 36.9 & 51.7 & 65.0 & 71.7 & 53.3 & 41.8 & 23.7 & 49.16 \\
            MobiLlama~\cite{mobillama} & 500M & 24.02& 1.25T& 51.1 & 53.4 & 70.9 & 76.4 & 55.7 & 46.0 & 26.6 & 54.30 \\
            OpenELM~\cite{openelm} & 450M & 24.04& 1.5T& 54.0 & 58.0 & 72.3 & 79.4 & 55.8 & 48.1 & 27.6 & 56.46 \\
            SmolLM~\cite{smollm} & 360M & 24.07& 600B& 53.5 & 56.8 & 71.5 & 84.2 & 55.4 & 63.8 & 36.0 & 60.17 \\
            \hline
            \textcolor{lightgray}{Qwen 1.5~\cite{qwen_1_5}} & \textcolor{lightgray}{500M} & \textcolor{lightgray}{24.02}& \textcolor{lightgray}{2.4T} & \textcolor{lightgray}{49.2} & \textcolor{lightgray}{55.7} & \textcolor{lightgray}{69.5} & \textcolor{lightgray}{82.5} & \textcolor{lightgray}{49.5} & \textcolor{lightgray}{52.3} & \textcolor{lightgray}{29.4} & \textcolor{lightgray}{55.44} \\
            \textcolor{lightgray}{Cerebras-GPT}~\cite{cerebras_cerebras_gpt_590m} & \textcolor{lightgray}{590M} & \textcolor{lightgray}{23.03}& \textcolor{lightgray}{371B}& \textcolor{lightgray}{32.3} & \textcolor{lightgray}{49.8} & \textcolor{lightgray}{62.8} & \textcolor{lightgray}{68.2} & \textcolor{lightgray}{59.2} & \textcolor{lightgray}{41.2} & \textcolor{lightgray}{23.5} & \textcolor{lightgray}{48.14} \\
            \hline
            \textbf{PhoneLM} & \textbf{500M} & 24.11& 1.1T& \textbf{54.0} & \textbf{57.9} & \textbf{73.3} & \textbf{85.1} & \textbf{60.7} & \textbf{60.4} & \textbf{31.6} & \textbf{60.43} \\
            \hline
        \end{tabular}
        \label{tab:performance-0.5B}
    }
    \quad
    \subfloat[1.5B]{
        \begin{tabular}{l|c|c|c|ccccc|cc|c}
            \hline
            \textbf{Name} & \textbf{Size} & \textbf{Date} & \makecell{\textbf{Training}\\\textbf{tokens}} & \textbf{HellaSwag} & \textbf{WinoGrande} & \textbf{PIQA} & \textbf{SciQ} & \textbf{BoolQ} & \makecell{\textbf{ARC}\\\textbf{Easy}} & \makecell{\textbf{ARC}\\\textbf{Challenge}} & \textbf{Average} \\
            \hline
            Pythia~\cite{eleutherai_pythia_1.4b} & 1.4B & 23.03& 207B& 52.0 & 57.2 & 71.1 & 79.2 & 63.2 & 53.9 & 28.3 & 57.84 \\
            OPT~\cite{facebook_opt_1.3b} & 1.3B & 22.05&180B& 53.7 & 59.0 & 71.0 & 78.1 & 57.2 & 51.3 & 28.0 & 56.90 \\
            BLOOM~\cite{bigscience_bloomz_1b1} & 1.1B & 22.11& 350B& 43.0 & 54.9 & 67.2 & 74.6 & 59.1 & 45.4 & 25.6 & 52.83 \\
            TinyLlama~\cite{tinyllama} & 1.1B & 23.12& 3B& 59.1 & 58.9 & 73.0 & 82.3 & 58.6 & 55.7 & 31.0 & 59.80 \\
            MobileLLaMA~\cite{mobilellama} & 1.4B & 23.12& 1.3T& 56.1 & 59.4 & 73.0 & 81.9 & 56.7 & 55.8 & 30.3 & 59.03 \\
            MobiLlama~\cite{mobillama} & 1B & 24.02& 1.25T& 62.2 & 59.3 & 74.8 & 82.8 & 60.3 & 56.4 & 31.7 & 61.07 \\
            OpenELM~\cite{openelm} & 1.1B & 24.04& 1.5T& 64.8 & 61.7 & 75.6 & 83.6 & 63.6 & 55.4 & 32.3 & 62.43 \\
            DCLM~\cite{dclm} & 1.4B & 24.08& 4.3T& 53.6 & 66.3 & 77.0 & 94.0 & 71.4 & 74.8 & 41.2 & 68.33 \\
            SmolLM~\cite{smollm} & 1.7B & 24.07& 1T& 49.6 & 60.9 & 75.8 & 93.2 & 66.0 & 76.4 & 43.5 & 66.49 \\
            \hline
            \textcolor{lightgray}{Qwen 1.5~\cite{qwen_1_5}} & \textcolor{lightgray}{1.8B} & \textcolor{lightgray}{24.02}& \textcolor{lightgray}{2.4T}& \textcolor{lightgray}{60.9} & \textcolor{lightgray}{60.5} & \textcolor{lightgray}{74.2} & \textcolor{lightgray}{89.4} & \textcolor{lightgray}{66.5} & \textcolor{lightgray}{59.1} & \textcolor{lightgray}{34.7} & \textcolor{lightgray}{63.61} \\
            \textcolor{lightgray}{Galactica~\cite{facebook_galactica_1.3b}} & \textcolor{lightgray}{1.3B} & \textcolor{lightgray}{22.11}& \textcolor{lightgray}{106B}& \textcolor{lightgray}{41.0} & \textcolor{lightgray}{54.4} & \textcolor{lightgray}{63.8} & \textcolor{lightgray}{87.7} & \textcolor{lightgray}{62.0} & \textcolor{lightgray}{58.6} & \textcolor{lightgray}{30.5} & \textcolor{lightgray}{56.86} \\
            \textcolor{lightgray}{StableLM 2~\cite{stabilityai_stablelm_2_zephyr}} & \textcolor{lightgray}{1.6B} & \textcolor{lightgray}{24.01}&\textcolor{lightgray}{2T} & \textcolor{lightgray}{68.8} & \textcolor{lightgray}{64.1} & \textcolor{lightgray}{75.1} & \textcolor{lightgray}{76.9} & \textcolor{lightgray}{80.0} & \textcolor{lightgray}{60.3} & \textcolor{lightgray}{39.2} & \textcolor{lightgray}{66.34} \\
            \textcolor{lightgray}{Cerebras-GPT~\cite{cerebras_cerebras_gpt_1.3b}} & \textcolor{lightgray}{1.3B} & \textcolor{lightgray}{23.03}& \textcolor{lightgray}{371B} & \textcolor{lightgray}{38.4} & \textcolor{lightgray}{51.9} & \textcolor{lightgray}{66.8} & \textcolor{lightgray}{73.0} & \textcolor{lightgray}{59.3} & \textcolor{lightgray}{45.8} & \textcolor{lightgray}{25.3} & \textcolor{lightgray}{51.50} \\
            \textcolor{lightgray}{MiniCPM~\cite{minicpm}} & \textcolor{lightgray}{1B} & \textcolor{lightgray}{24.04}&\textcolor{lightgray}{1.2T} & \textcolor{lightgray}{67.5} & \textcolor{lightgray}{63.7} & \textcolor{lightgray}{75.1} & \textcolor{lightgray}{91.0} & \textcolor{lightgray}{70.5} & \textcolor{lightgray}{62.9} & \textcolor{lightgray}{38.1} & \textcolor{lightgray}{66.97} \\
            \textcolor{lightgray}{MiniCPM~\cite{minicpm}} & \textcolor{lightgray}{2B} & \textcolor{lightgray}{24.04}& \textcolor{lightgray}{1.2T}& \textcolor{lightgray}{67.2} & \textcolor{lightgray}{63.9} & \textcolor{lightgray}{76.1} & \textcolor{lightgray}{92.5} & \textcolor{lightgray}{74.6} & \textcolor{lightgray}{69.0} & \textcolor{lightgray}{42.7} & \textcolor{lightgray}{69.43} \\
            \textcolor{lightgray}{Gemma~\cite{gemma} } & \textcolor{lightgray}{2B} & \textcolor{lightgray}{24.02}& \textcolor{lightgray}{3T}& \textcolor{lightgray}{71.4} & \textcolor{lightgray}{65.2} & \textcolor{lightgray}{78.4} & \textcolor{lightgray}{91.4} & \textcolor{lightgray}{69.9} & \textcolor{lightgray}{72.3} & \textcolor{lightgray}{42.0} & \textcolor{lightgray}{70.09} \\
            \textcolor{lightgray}{Gemma 2~\cite{gemma_2}} & \textcolor{lightgray}{2B} & \textcolor{lightgray}{24.07}& \textcolor{lightgray}{2T}& \textcolor{lightgray}{55.0} & \textcolor{lightgray}{68.7} & \textcolor{lightgray}{78.7} & \textcolor{lightgray}{96.0} & \textcolor{lightgray}{73.6} & \textcolor{lightgray}{80.3} & \textcolor{lightgray}{46.9} & \textcolor{lightgray}{71.31} \\
            \hline
            \textbf{PhoneLM} & \textbf{1.5B} &24.11 &1.5T & \textbf{66.9} & \textbf{63.0} & \textbf{77.3} & \textbf{88.8} & \textbf{65.5} & \textbf{69.7} & \textbf{39.9} & \textbf{67.31} \\
            \hline
        \end{tabular}
        \label{tab:performance-1.5B}
    }
    \caption{Benchmark Score of \llm. Models in grey color are trained on proprietary datasets.
    }
    \label{tab:performance}
\end{table*}
\begin{figure*}[t] 
	\centering  
	\subfloat[0.5B]{
        \label{fig:experimental-step-0.5}
		\includegraphics[width=0.5\linewidth]{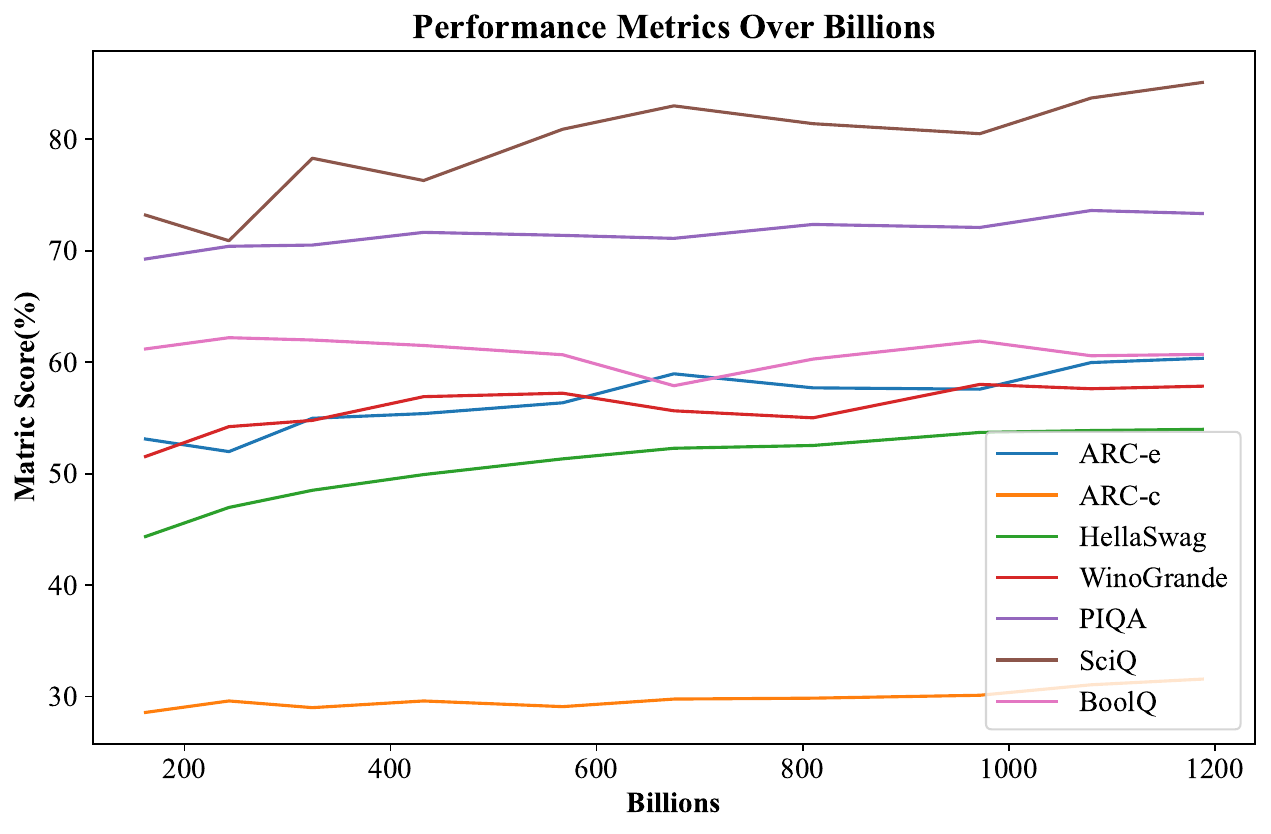}}
	\subfloat[1.5B]{
		\label{fig:experimental-step-1.5B}
		\includegraphics[width=0.5\linewidth]{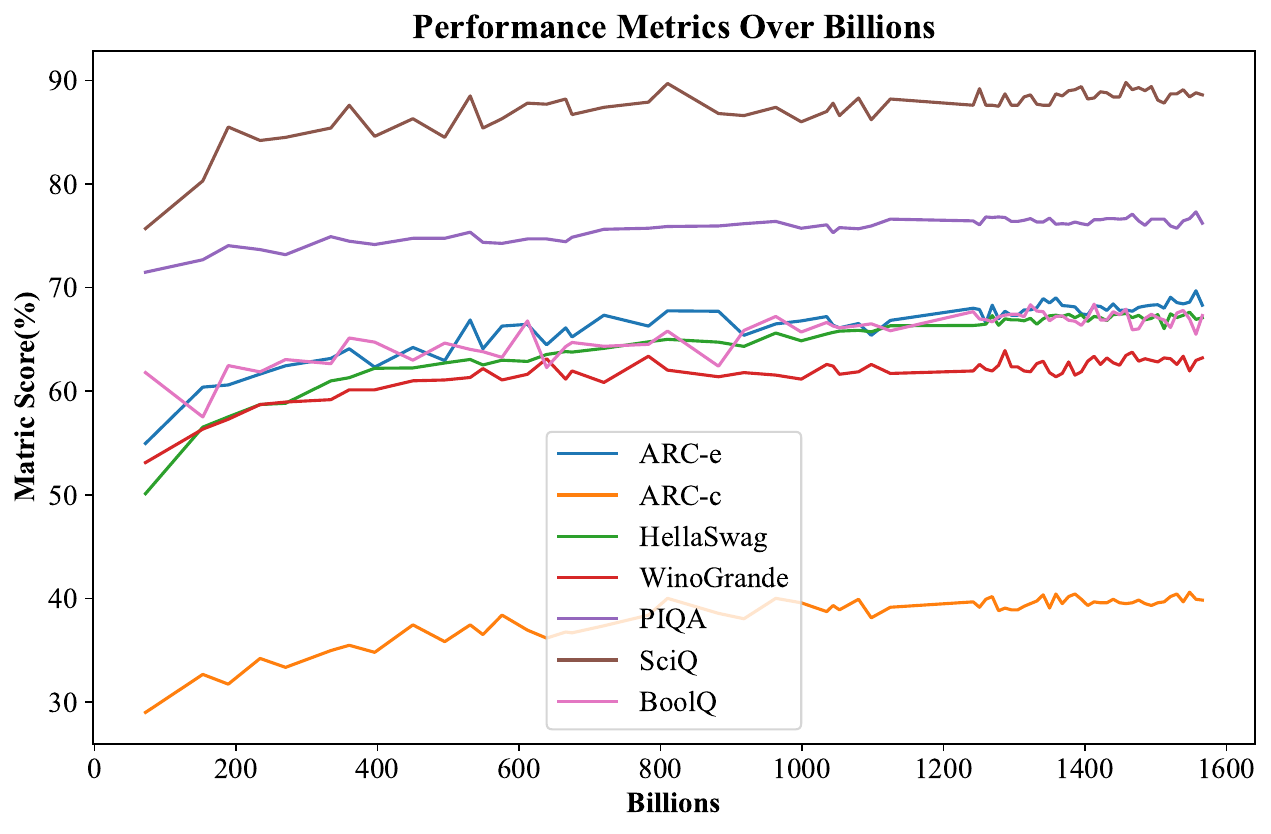}}
	\caption{\llm's performance across training iterations on standard zero-shot tasks}
	\label{fig:experimental-step}
\end{figure*}


We evaluate \llm on a wide range of commonsense reasoning and problem-solving tasks and compare it to several existing open-source language models with similar model sizes.

\subsection{Baselines and Tasks}
We compare \llm family models to several existing open-source language models with similar model sizes.
For \llm-0.5B, the main comparisons are made among models with fully open-source datasets, code, and weights, including Pythia-410M, OPT-350M, BLOOM-560M, MobiLlama-0.5B, OpenELM-450M, and SmolLM-360M. 
At the same time, we also compared with some models that only open-source the weights, such as Qwen1.5-0.5B, LaMini-GPT-774M, and Cerebras-GPT-590M.
For \llm-1.5B, the models with fully open-source code and weights include Pythia-1.4B, OPT-1.3B, BLOOM-1.1B, TinyLlama-1.1B, MobileLLaMA-1.4B, MobiLlama-1B, OpenELM-1.1B, DCLM-1B, SmolLM-1.7B. The models that only open-source the weights include Qwen1.5-1.8B, StableLM2-1.6B, MiniCPM-2B, Gemma2-2B, etc. 

To understand the ability of \llm, 
We used 7 datasets across two domains to evaluate the SLM performance.
\begin{itemize}
\item \textbf{Commonsense Reasoning Datasets}:
    \begin{itemize}
    \item \textbf{HellaSwag}~\cite{zellers2019hellaswag}: Tests narrative understanding through plausible sentence completion.
    \item \textbf{Winogrande}~\cite{sakaguchi2020winogrande}: Evaluates pronoun ambiguity resolution using commonsense reasoning.
    \item \textbf{PIQA}~\cite{bisk2020piqa}: Focuses on physical commonsense reasoning and object interactions.
    \item \textbf{SciQ}~\cite{welbl2017crowdsourcing}: a dataset of 13.7K multiple choice science exam questions.
    \item \textbf{BoolQ}~\cite{clark2019boolq}: Tests commonsense and factual reasoning with yes/no questions.
    \end{itemize}
\item \textbf{Problem Solving Datasets}:
    \begin{itemize}
    \item \textbf{ARC Easy}~\cite{clark2018think}: Contains simple science questions testing general knowledge and reasoning.
    \item \textbf{ARC Challenge}~\cite{clark2018think}: Presents complex science exam questions requiring knowledge integration.
    \end{itemize}
\end{itemize}

We adopt the benchmark \textit{lm\_eval}~\cite{lmevaluation} to evaluate the models. 
We evaluate the models after stable training stage.
We use \textit{accuracy} as the primary evaluation metric.
Accuracy measures the proportion of correct predictions to total examples.
For commonsense reasoning and problem solving tasks, accuracy evaluates the model's ability to select correct options or provide accurate solutions.
Following previous practice, the models are evaluated in a zero-shot setting on these tasks.
We notice that \llm outperforms baselines on many of the tasks and obtains the highest averaged scores for most open-source models.
We also evaluate the models on other following tasks, which contains the following tasks:
SocialIQA, TruthfulQA, MMLU, CMMLU and C-Eval.

\subsection{Capability}\label{sec:pretraining-results}
The capability for 7 standard zero-shot tasks of \llm are presented in table~\ref{tab:performance}. 
It can be seen from table~\ref{tab:performance}(a) that \llm-0.5B achieves the highest average accuracy on these 7 tasks.
Except for the two tasks of ARC-e and ARC-c, where \llm-0.5B performs lower than SmolLM, \llm-0.5B demonstrates the strongest performance on other tasks among models with similar parameter counts.
For \llm-1.5B, which is shown in table~\ref{tab:performance}(b), it performs better than other open-source models on most tasks.
Combining all the tasks, it can be seen that \llm performs better than other models with the same number of parameters in Commonsense Reasoning tasks (e.g., ARC-c, HellaSwag, and PIQA), and Problem Solving tasks (e.g., ARC-Easy and ARC-Challenge).

In figure~\ref{fig:experimental-step}, the accuracy of \llm-0.5B and \llm-1.5B are plotted against training iterations for 7 standard zero-shot tasks. 
We observe an overall increase in accuracy with longer training durations across most tasks. 

\subsection{Instruction and Function Call}
\textbf{Instruction following.}
We have attached examples of \llm-1.5B-Instruction in several scenarios, including "Reasoning", "Knowledge", "Programming and Logic Building", "Innovative Thinking", "Translation", and "Creativity and Imagination" in appendix~\ref{sec:appendix-instruct}.

\textbf{Function call.}
We first define two metrics to evaluate the performance of the function calls: \textit{Accuracy} and \textit{Soft Accuracy}.
\begin{itemize}
    \item \textit{Accuracy:} A sample contains a user query and its corresponding ground-truth function calls. A sample is considered correct only if the model generates all function calls with both correct functions and parameters. Accuracy is defined as the ratio of correctly predicted samples to the total number of samples.
    \item \textit{Soft Accuracy:} To provide a more fine-grained evaluation when the model generates partially correct results (i.e., correct functions with partially correct parameters), we define soft accuracy. For each function call, a score is calculated as the ratio of correctly predicted parameters to the total number of parameters. Soft accuracy is then computed as the average of these scores across all function calls.
\end{itemize}

To evaluate the model's inherent function calling capabilities, we crafted comprehensive prompts to guide the chat model in executing function calls. 
Subsequently, we fine-tuned the \llm following the methodology described in Section~\ref{finetune}.  We tested some common models on the DroidCall test set.
The results are presented in Table~\ref{tab:finetune}.

\begin{table}[!ht]
    \centering
    \small
    \begin{tabular}{l|r|r}
    \hline
        \textbf{Model} & \textbf{Accuracy} & \textbf{Soft Accuracy} \\ 
    \hline
        Qwen2.5-Coder-1.5B & 50.0 & 63.5 \\ 
        Qwen2.5-1.5B-Instruct & 58.5 & 75.3 \\ 
        Phi-3.5-mini-instruct & 62.0 & 77.7 \\ 
        MiniCPM3-4B & 70.0 & 85.7 \\ 
        Gemma-2-2b-it & 56.5 & 75.8 \\ 
        TinyLlama-1.1B-Chat-v1.0 & 18.0 & 18.7 \\ 
        Llama-3.2-1B-Instruct & 36.0 & 43.8 \\ 
        Llama-3.2-3B-Instruct & 47.5 & 57.9 \\ 
        GPT-4o-mini & 71.0 & 86.1 \\  \hline
        \textbf{\llm-1.5B-Instruct} & \textbf{17.5} & \textbf{17.8}\\ 
        \textbf{\llm-1.5B-Call} & \textbf{75.0} & \textbf{86.1} \\ 
    \hline
    \end{tabular}
    \caption{Performance comparison of different models on the DroidCall test set. }
    \label{tab:finetune}
\end{table}

\subsection{On-device Runtime Cost}
\begin{figure*}[ht]
    \vskip 0.2in
    \begin{center}
    \centerline{\includegraphics[width=2\columnwidth]{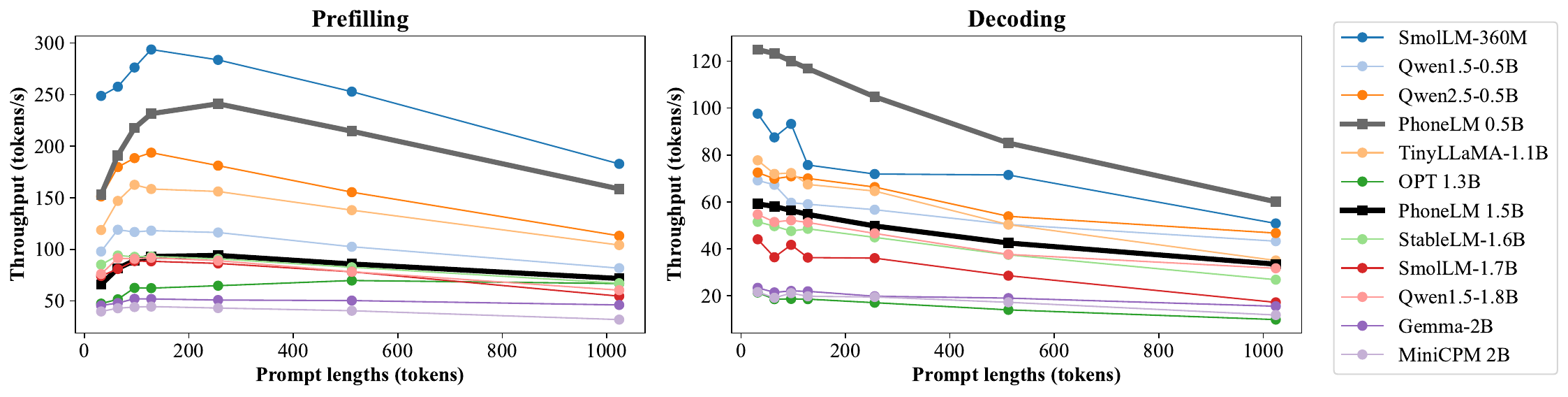}}
    \caption{\llm's throughput.}
    \label{fig:benckmark-speed}
    \end{center}
    \vskip -0.2in
\end{figure*}

\textbf{Hardware and framework.}
To benchmark \llm models on the smartphone, we used a Xiaomi 14 with a Qualcomm Snapdragon 8 Gen 3 system-on-chip (SoC) and 16GiB of RAM, running Android 14.
We set the smartphone as a performance mode to ensure stable benchmark results.
We ported the code and weights of \llm to \textit{mllm}~\cite{mllm}.
We conducted experiments in both CPU and NPU usage scenarios.
For the CPU-only benchmark, we used four threads for computation, deploying them on the four performance cores of the CPU.
The weights of the linear and embedding layers in \textit{mllm} were quantized to 4-bit, while the weights of the RMSNorm layer were kept at fp32. 
During runtime, the activation value is in fp32 format. 
To accelerate inference, 4-bit weights were calculated in units of 4$\times$4 groups.
When conducting tests using the NPU, we adopted Qualcomm's QNN~\cite{qualcomm-qnn} framework and used the methods mentioned in mllm-NPU~\cite{xu2024empowering}.  

\textbf{Evaluation.}
We provide two separate measurements for token throughput (measured in tokens processed per second): (1) prompt processing (prefilling), and (2) token generation(decoding). 
In the benchmark experiments of the model, we set different prompt lengths ranging from 32 to 1024 tokens and generate 100 tokens in an autoregressive manner to measure the throughput in the prefilling stage and the decoding stage. 
We conduct 5 experiments for each model and take the average result. 
We use key-value caching in all experiments.

\textbf{Results.}
Figure~\ref{fig:benckmark-speed} shows the benchmarking results on Xiaomi 14(CPU).
The prefilling throughput of \llm-0.5B is higher than that of all 0.5B and larger models except SmolLM-360M, and its decoding throughput is higher than that of all models.
The prefilling throughput and decoding throughput of \llm-1.5B are greater than those of all 1.3B and larger models.
Overall, as the prompt length increases, the throughput decreases. This is because the increase in the prompt length leads to an increase in the computational load of self-attention.
We compared the throughputs of different models at a prompt length of 64 with their average metrics and plotted them in Figure~\ref{fig:benckmark-throughput-matrics}. 
In the figure, the models positioned closer to the upper right corner have better performance and speed. 
It can be seen that the two sizes of \llm are better than the models with similar parameter counts. 
The prefilling speed of \llm-1.5B even achieves 654 tokens/second on Xiaomi 14 NPU, which is faster than Qwen 2.5 1.5B (602 tokens/second).




\subsection{An End-to-end Android Demo}
We also have an end-to-end Android demo application for \llm-1.5B based on \textit{mllm}.
This demo contains two invocations: chat and Android intent invocation.
The screenshots of this application are shown in figure~\ref{fig:experimental-innvaction}.
Figure~\ref{fig:experimental-innvaction}(a) shows an example of a user having a conversation with an Android application with \llm-1.5B-Instruct built in.
Figure~\ref{fig:experimental-innvaction}(b) shows the Android intent invocation ability of the \llm-1.5B-Call model. In this example, after understanding the user's "Wake me up at 8:00", the model uses the Android alarm-setting Intent to set an alarm for 8 o'clock.  
\section{Conclusions}

This work presents \llm, an efficient, capable, and fully open-sourced small language family.
\llm is built atop a unique principle: searching for a runtime-efficient transformer architecture ahead of pre-training.
We also release an end-to-end demo using \llm for intent invocations on Android OS in a fast and accurate performance.
The goal of \llm is to advance the development and research on small language models towards more practical on-device deployment.




\appendix
\section{Setting of 100M and 200M models}
\label{sec:appendix-100200-setting}

\begin{table*}[t]
    \centering
	\subfloat[100M]{
        \begin{tabular}{c|c|cccccc|c|cc}
            \hline
            \textbf{ID} &  \textbf{size(M)} &\textbf{hidden} & \textbf{intermediate}  & \textbf{layers}  & \textbf{activation} & \makecell{\textbf{q}\\\textbf{heads}} & \makecell{\textbf{kv}\\\textbf{heads}} & \textbf{loss} & \makecell{\textbf{prefilling}\\\textbf{(tokens/s)}} &  \makecell{\textbf{decoding}\\\textbf{(tokens/s)}} \\
            \hline
            1 & 106.73 & 1280 & 2096 & 3 & relu & 16 & 16 & 3.76 & 916.70 & 455.32 \\
            2 & 106.73 & 1280 & 2096 & 3 & silu & 16 & 16 & 3.81 & 877.19 & 424.08 \\
            \hline
            3 & 101.42 & 768 & 2046 & 9 & relu & 16 & 16 & 3.70 & 742.85 & 258.56 \\
            4 & 101.42 & 768 & 2046 & 9 & relu & 4 & 4 & 3.67 & 784.94 & 266.68 \\
            5 & 101.42 & 768 & 2046 & 9 & relu & 16 & 4 & 3.66 & 871.94 & 260.37 \\
            6 & 101.42 & 768 & 2046 & 9 & silu & 16 & 16 & 3.69 & 788.95 & 260.03 \\
            7 & 101.42 & 768 & 2046 & 9 & silu & 4 & 4 & 3.66 & 773.27 & 255.42 \\
            8 & 101.42 & 768 & 2046 & 9 & silu & 16 & 4 & 3.65 & 853.46 & 252.71 \\
            \hline
            9 & 99.54 & 704 & 1856 & 11 & relu & 16 & 16 & 3.65 & 720.98 & 228.11 \\
            10 & 99.54 & 704 & 1856 & 11 & silu & 16 & 16 & 3.64 & 753.61 & 228.03 \\
            \hline
            11 & 100.00 & 576 & 1536 & 18 & relu & 16 & 16 & 3.68 & 601.56 & 154.59 \\
            12 & 100.00 & 576 & 1536 & 18 & relu & 4 & 4 & 3.59 & 652.11 & 164.05 \\
            13 & 100.00 & 576 & 1536 & 18 & relu & 16 & 4 & 3.66 & 705.54 & 153.85 \\
            14 & 100.00 & 576 & 1536 & 18 & silu & 16 & 16 & 3.67 & 614.41 & 151.98 \\
            15 & 100.00 & 576 & 1536 & 18 & silu & 4 & 4 & 3.58 & 640.13 & 160.48 \\
            16 & 100.00 & 576 & 1536 & 18 & silu & 16 & 4 & 3.65 & 691.67 & 150.15 \\
            \hline
            17 & 101.06 & 448 & 1184 & 33 & relu & 16 & 16 & 3.68 & 469.89 & 89.48 \\
            18 & 101.06 & 448 & 1184 & 33 & silu & 16 & 16 & 3.67 & 481.58 & 87.70 \\
            \hline
        \end{tabular}
    }
	\quad 
	\subfloat[200M]{
        \begin{tabular}{c|c|cccccc|c|cc}
            \hline
            \textbf{ID} &  \textbf{size(M)} &\textbf{hidden} & \textbf{intermediate}  & \textbf{layers}  & \textbf{activation} & \makecell{\textbf{q}\\\textbf{heads}} & \makecell{\textbf{kv}\\\textbf{heads}} & \textbf{loss} & \makecell{\textbf{prefilling}\\\textbf{(tokens/s)}} &  \makecell{\textbf{decoding}\\\textbf{(tokens/s)}} \\
            \hline
            1 & 201.32 & 2048 & 5460 & 2 & relu & 16 & 16 & 4.21 & 726.44 & 430.06 \\
            2 & 201.32 & 2048 & 5460 & 2 & silu & 16 & 16 & 4.21 & 552.06 & 325.93 \\
            \hline
            3 & 188.76 & 1536 & 4096 & 4 & relu & 16 & 16 & 3.94 & 706.14 & 391.36 \\
            4 & 188.76 & 1536 & 4096 & 4 & silu & 16 & 16 & 3.91 & 683.97 & 351.09 \\
            \hline
            5 & 199.78 & 1024 & 2688 & 12 & relu & 16 & 16 & 3.89 & 559.80 & 225.88 \\
            6 & 199.78 & 1024 & 2688 & 12 & relu & 4 & 4 & 3.87 & 533.00 & 222.27 \\
            7 & 199.78 & 1024 & 2688 & 12 & relu & 16 & 4 & 3.89 & 546.76 & 215.04 \\
            8 & 199.78 & 1024 & 2688 & 12 & silu & 16 & 16 & 3.85 & 461.42 & 178.95 \\
            9 & 199.78 & 1024 & 2688 & 12 & silu & 4 & 4 & 3.86 & 427.38 & 162.85 \\
            10 & 199.78 & 1024 & 2688 & 12 & silu & 16 & 4 & 3.86 & 412.81 & 160.71 \\
            \hline
            11 & 182.20 & 704 & 1856 & 25 & relu & 16 & 16 & 4.02 & 489.62 & 144.05 \\
            12 & 182.20 & 704 & 1856 & 25 & relu & 4 & 4 & 3.95 & 505.01 & 139.14 \\
            13 & 182.20 & 704 & 1856 & 25 & relu & 16 & 4 & 3.98 & 554.88 & 131.29 \\
            14 & 182.20 & 704 & 1856 & 25 & silu & 16 & 16 & 3.98 & 487.49 & 124.17 \\
            15 & 182.20 & 704 & 1856 & 25 & silu & 4 & 4 & 3.95 & 391.94 & 103.51 \\
            16 & 182.20 & 704 & 1856 & 25 & silu & 16 & 4 & 3.94 & 448.85 & 98.58 \\
            \hline
            17 & 187.61 & 576 & 1536 & 40 & relu & 16 & 16 & 4.11 & 430.52 & 119.42 \\
            18 & 187.61 & 576 & 1536 & 40 & silu & 16 & 16 & 4.13 & 407.08 & 88.21 \\
            \hline
        \end{tabular}
    }
    \caption{100M and 200M models' setting}
    \label{tab:motivation-100_200M-matrics-show}
\end{table*}

Tested the speed and performance of 100M and 200M models, training on data with 20 billion tokens. 
The settings are shown in the table~\ref{tab:motivation-100_200M-matrics-show}.




\section{Training Dataets}
\label{sec:appendix-dataset}

DCLM-baseline~\cite{li2024datacomplm} is a 4T token / 3B document pretraining dataset that achieves strong performance on language model benchmarks.\llm only uses a maximum of 1.5T among it.
The code is publiced in \url{https://huggingface.co/datasets/mlfoundations/dclm-baseline-1.0-parquet}.

StarCoderData~\cite{li2023starcoder} contains 783GB of code in 86 programming languages, and includes 54GB GitHub Issues + 13GB Jupyter notebooks in scripts and text-code pairs, and 32GB of GitHub commits, which is approximately 250 Billion tokens.
The code is publiced in \url{https://huggingface.co/datasets/bigcode/starcoderdata}.

OpenWebMath~\cite{paster2023openwebmath} is a dataset containing the majority of the high-quality, mathematical text from the internet. It is filtered and extracted from over 200B HTML files on Common Crawl down to a set of 6.3 million documents containing a total of 14.7B tokens.
The code is publiced in \url{https://huggingface.co/datasets/open-web-math/open-web-math}.

Dolma~\cite{dolma} is a dataset of 3 trillion tokens from a diverse mix of web content, academic publications, code, books, and encyclopedic materials.
The code is publiced in \url{https://huggingface.co/datasets/allenai/dolma}.

APIGen~\cite{liu2024apigen} contains 60,000 data collected by APIGen, an automated data generation pipeline designed to produce verifiable high-quality datasets for function-calling applications.
The code is publiced in \url{https://huggingface.co/datasets/Salesforce/xlam-function-calling-60k}.

The Stack Smol~\cite{kocetkov2022stack} is a small subset of the-stack dataset, each programming language has 10,000 random samples from the original dataset. 
The code is publiced in \url{https://huggingface.co/datasets/bigcode/the-stack-smol}.

UltraChat~\cite{ding2023enhancing} is an open-source, large-scale, and multi-round dialogue data powered by Turbo APIs.
The code is publiced in \url{https://huggingface.co/datasets/stingning/ultrachat}.

MathInstruct~\cite{yue2023mammoth} is a meticulously curated instruction tuning dataset that is lightweight yet generalizable. 
The code is publiced in \url{https://huggingface.co/datasets/TIGER-Lab/MathInstruct}.

OpenAssistant 2~\cite{kopf2024openassistant} contains message trees. Each message tree has an initial prompt message as the root node, which can have multiple child messages as replies, and these child messages can have multiple replies.
The code is publiced in \url{https://huggingface.co/datasets/OpenAssistant/oasst2}.

OpenHermes~\cite{openhermes} dataset is composed of 242,000 entries of primarily GPT-4 generated data, from open datasets across the AI landscape.
The code is publiced in \url{https://huggingface.co/datasets/teknium/openhermes}.

CommitPackFT~\cite{muennighoff2023octopack} is a 2GB filtered version of CommitPack to contain only high-quality commit messages that resemble natural language instructions.
The code is publiced in \url{https://huggingface.co/datasets/bigcode/commitpackft}.

OSS Instruct dataset~\cite{MagicoderOSSInstruct75K} is generated by gpt-3.5-turbo-1106 developed by OpenAI.
The code is publiced in \url{https://huggingface.co/datasets/ise-uiuc/Magicoder-OSS-Instruct-75K}.

SlimOrca~\cite{SlimOrca} release provides an efficient means of reaching performance on-par with using larger slices of our data, while only including ~500k GPT-4 completions.
The code is publiced in \url{https://huggingface.co/datasets/Open-Orca/SlimOrca}.

\newcommand{\OutputBoxxx}[2]{%
    \newtcolorbox{#1}{
      colback=white,
      colframe={black!65},
      title={\textbf{#2}},
      fonttitle=\bfseries,
      boxrule=0.5mm,
      arc=3mm,
      boxsep=5pt,
      left=0pt,
      right=6pt,
      top=8pt,
      bottom=8pt,
    }
}
\section{Instruct Following Examples}
\label{sec:appendix-instruct}

\OutputBoxxx{OutputBoxxxA}{Reasoning:}
\begin{OutputBoxxxA}
\begin{itemize}[left=0pt, label={},noitemsep,topsep=0pt]
  \item \textbf{\textcolor{blue}{Prompt:}} What are the key benefits of practicing mindfulness meditation?
  \begin{itemize}[left=0pt,label={},noitemsep,topsep=0pt]
  \item
  \end{itemize}

  \item \textbf{\textcolor{orange}{PhoneLM-1.5B-Instruct:}}
  \begin{itemize}[left=0pt,label={},noitemsep,topsep=0pt]
    \item   There are many benefits of practicing mindfulness meditation, including:
    \item 
    \item   1. Reduced stress and anxiety
    \item   2. Improved focus and concentration
    \item   3. Increased self-awareness and emotional regulation
    \item   4. Improved sleep quality
    \item   5. Reduced symptoms of depression
    \item   6. Increased feelings of well-being and happiness
    \item   7. Improved immune system function
    \item   8. Reduced symptoms of chronic pain
    \item   9. Improved relationships and communication skills
    \item   10. Increased feelings of compassion and empathy.
\end{itemize}

\end{itemize}
\end{OutputBoxxxA}

\OutputBoxxx{OutputBoxxxB}{Knowledge:}
\begin{OutputBoxxxB}
\begin{itemize}[left=0pt, label={},noitemsep,topsep=0pt]
  \item \textbf{\textcolor{blue}{Prompt:}} What are some ways to reduce the carbon footprint of your home?
  \begin{itemize}[left=0pt,label={},noitemsep,topsep=0pt]
  \item
  \end{itemize}

  \item \textbf{\textcolor{orange}{PhoneLM-1.5B-Instruct:}}
  \begin{itemize}[left=0pt,label={},noitemsep,topsep=0pt]
  \item There are several ways to reduce the carbon footprint of your home. Here are some suggestions:
  \item 
  \item 1. Install energy-efficient appliances and lighting.
  \item 2. Use low-flow showerheads and faucets.
  \item 3. Install a programmable thermostat to regulate heating and cooling.
  \item 4. Use natural light and ventilation instead of artificial lighting and air conditioning.
  \item 5. Plant trees or install shading devices to block the sun's heat.
  \item 6. Use eco-friendly cleaning products and reduce the use of disposable products.
  \item 7. Reduce water usage by fixing leaks and installing low-flow fixtures.
  \item 8. Use renewable energy sources such as solar panels or wind turbines.
  \item 9. Reduce waste by recycling, composting, and using reusable products.
  \item 10. Choose sustainable building materials and design your home to be energy-efficient.
\end{itemize}

\end{itemize}
\end{OutputBoxxxB}

\OutputBoxxx{OutputBoxxxC}{Programming and Logic Building:}
\begin{OutputBoxxxC}
\begin{itemize}[left=0pt, label={},noitemsep,topsep=0pt]
  \item \textbf{\textcolor{blue}{Prompt:}} Generate a Python code snippet that implements a function to calculate the Fibonacci sequence using recursion. Bonus points if you can make it more efficient using memoization.
  \begin{itemize}[left=0pt,label={},noitemsep,topsep=0pt]
  \item
  \end{itemize}

  \item \textbf{\textcolor{orange}{PhoneLM-1.5B-Instruct:}}
  \begin{itemize}[left=0pt,label={},noitemsep,topsep=0pt]
  \item \verb|def fibonacci(n):|
  \item \quad \verb|if n <= 1:|
  \item \quad \quad \verb|return n|
  \item \quad \verb|else:|
  \item \quad \quad \verb|return fibonacci(n-1) + |
  \item \quad \quad \quad \quad \quad \quad \verb|fibonacci(n-2)| \\
  \item 
  \item \verb|# Test the function|
  \item \verb|print(fibonacci(0)) # 0|
  \item \verb|print(fibonacci(1)) # 1|
  \item \verb|print(fibonacci(2)) # 1|
  \item \verb|print(fibonacci(3)) # 2|
  \item \verb|print(fibonacci(4)) # 3|
  \item \verb|print(fibonacci(5)) # 5|
  \item \verb|print(fibonacci(6)) # 8|
  \item \verb|print(fibonacci(7)) # 13|
  \item \verb|print(fibonacci(8)) # 21|
  \item \verb|print(fibonacci(9)) # 34|
  \item \verb|print(fibonacci(10)) # 55|

\end{itemize}

\end{itemize}
\end{OutputBoxxxC}

\OutputBoxxx{OutputBoxxxD}{Innovative Thinking:}
\begin{OutputBoxxxD}
\begin{itemize}[left=0pt, label={},noitemsep,topsep=0pt]
  \item \textbf{\textcolor{blue}{Prompt:}} Construct a sentence that uses every letter of the alphabet at least once.
  \begin{itemize}[left=0pt,label={},noitemsep,topsep=0pt]
  \item
  \end{itemize}
  
  \item \textbf{\textcolor{orange}{PhoneLM-1.5B-Instruct:}}
  \begin{itemize}[left=0pt,label={},noitemsep,topsep=0pt]
  \item Able was I ere I saw Elba, but ere I saw Elba, I heard a rumor that would change my life forever.
\end{itemize}
\end{itemize}
\end{OutputBoxxxD}

\OutputBoxxx{OutputBoxxxE}{Translation:}
\begin{OutputBoxxxE}
\begin{itemize}[left=0pt, label={},noitemsep,topsep=0pt]
  \item \textbf{\textcolor{blue}{Prompt:}} Translate this sentence into French: ”Ming due to the circumstances, the knower system according to the matter.”
  \begin{itemize}[left=0pt,label={},noitemsep,topsep=0pt]
  \item
  \end{itemize}
  
  \item \textbf{\textcolor{orange}{PhoneLM-1.5B-Instruct:}}
  \begin{itemize}[left=0pt,label={},noitemsep,topsep=0pt]
  \item "Ming, parce qu'il y a des circonstances, le système d'apprentissage selon le matériel."
\end{itemize}
\end{itemize}
\end{OutputBoxxxE}

\OutputBoxxx{OutputBoxxxF}{Creativity and Imagination:}
\begin{OutputBoxxxF}
\begin{itemize}[left=0pt, label={},noitemsep,topsep=0pt]
  \item \textbf{\textcolor{blue}{Prompt:}} Write a role-play with a tourist and a travel guide for a finctional planet. 
  \begin{itemize}[left=0pt,label={},noitemsep,topsep=0pt]
  \item
  \end{itemize}
  
  \item \textbf{\textcolor{orange}{PhoneLM-1.5B-Instruct:}}
  \begin{itemize}[left=0pt,label={},noitemsep,topsep=0pt]
    \item Tourist: Hi, I'm looking for a planet that's not too crowded. Can you recommend one?
    \item 
    \item Travel Guide: Sure, I can recommend a few. One of the most popular planets is Zorion. It's a bit more remote, but it's still accessible.
    \item 
    \item Tourist: That sounds great. What's the best way to get there?
    \item 
    \item Travel Guide: You can take a shuttle from the nearest spaceport. It's about a 3-hour journey.
    \item 
    \item Tourist: That sounds doable. What's the atmosphere like on Zorion?
    \item 
    \item Travel Guide: It's a bit different from Earth. The atmosphere is mostly nitrogen and oxygen, but there's also a lot of carbon dioxide. It's a bit more humid too.
    \item 
    \item Tourist: That's interesting. What's the terrain like?
    \item 
    \item Travel Guide: It's mostly rocky and barren, but there are some forests and lakes. It's a bit like Mars, but with more vegetation.
    \item 
    \item Tourist: That sounds like a unique experience. Are there any popular tourist attractions on Zorion?
    \item 
    \item Travel Guide: Yes, there are a few. One of the most popular is the Zorion Museum, which showcases the planet's history and culture. Another is the Zorion Observatory, which is a great place to see the stars.
    \item 
    \item Tourist: That sounds amazing. Is there anything else I should know before I go?
    \item 
    \item Travel Guide: Just be prepared for the weather. It can get pretty hot and humid, so bring a lot of water and sunscreen. And don't forget your passport, as Zorion is a popular tourist destination.
    \item 
    \item Tourist: Thanks for the tips. I'm excited to visit Zorion.
\end{itemize}
\end{itemize}
\end{OutputBoxxxF}

  

\section{Function Calling fine-tuning details}\label{sec:appendix-function-calling}

We adopt a chat-based approach for function calling, where the requirements, use query, and function descriptions are encapsulated in the system prompt and user prompt sections, while the function calls are placed in the assistant output section. The prompt design for function calling fine-tuning thus focuses on structuring these three components: system prompt, user message, and assistant output. These components are subsequently formatted using a chat template, which is then utilized for model fine-tuning. These components are shown in \ref{lst:prompt}, in which \$function is the functions description information, which describes the function name, parameters, and other information, \$user\_query is the user input.

\begin{figure}[htbp]  
\begin{lstlisting}[caption={minimalist prompt of function calling}, label=lst:prompt]
System Prompt:

You are an expert in composing functions.

User message:

Here is a list of functions that you can invoke:
$functions
Now my query is: $user_query

Assistant output:

$result1 = func0(arg1="value1", arg2="value2", ...)
result2 = func1(arg1="value1", arg2=result1, ...)
...$
\end{lstlisting}
\end{figure}


\end{document}